\def\year{2024}\relax
\documentclass[letterpaper]{article} % DO NOT CHANGE THIS
\usepackage{aaai22}  % DO NOT CHANGE THIS
\usepackage{times}  % DO NOT CHANGE THIS
\usepackage{helvet}  % DO NOT CHANGE THIS
\usepackage{courier}  % DO NOT CHANGE THIS
\usepackage[hyphens]{url}  % DO NOT CHANGE THIS
\usepackage{graphicx} % DO NOT CHANGE THIS
\usepackage{natbib}  % DO NOT CHANGE THIS AND DO NOT ADD ANY OPTIONS TO IT
\usepackage{graphicx}
\usepackage{subcaption}
\usepackage{tabularx}
\usepackage{multirow}
\usepackage{lipsum}

\usepackage{caption} % DO NOT CHANGE THIS AND DO NOT ADD ANY OPTIONS TO IT
\DeclareCaptionStyle{ruled}{labelfont=normalfont,labelsep=colon,strut=off} % DO NOT CHANGE THIS
\usepackage{algorithm}
\usepackage{algorithmic}

\usepackage{newfloat}
\usepackage{listings}
\floatstyle{ruled}
\newfloat{listing}{tb}{lst}{}
\floatname{listing}{Listing}
\title{News Source Credibility Assessment: A Reddit Case Study}
\author {
    Arash Amini,\textsuperscript{\rm 1}
    Yigit Ege Bayiz,\textsuperscript{\rm 2}
    Ashwin Ram,\textsuperscript{\rm 2}
    Radu Marculescu,\textsuperscript{\rm 2} and 
    Ufuk Topcu \textsuperscript{\rm 1}
}

\affiliations {
    \textsuperscript{\rm 1} Oden Institute for Computational Engineering and Sciences, The University of Texas at Austin\\
     a.amini@utexas.edu, utopcu@utexas.edu\\
    \textsuperscript{\rm 2} Department of Electrical and Computer Engineering, The University of Texas at Austin\\
   egebayiz@utexas.edu, ashwin.ram@utexas.edu, radum@utexas.edu
}

\usepackage{bibentry}
\newcolumntype{b}{X}
\newcolumntype{s}{>{\hsize=.5\hsize}X}
\usepackage{amssymb}% http://ctan.org/pkg/amssymb
\usepackage{pifont}% http://ctan.org/pkg/pifont
\usepackage{amsmath}
\DeclareMathOperator*{\argmax}{argmax}

\usepackage{tikz}
\usetikzlibrary{shapes.geometric, arrows}

\tikzstyle{startstop} = [rectangle, rounded corners, minimum width=3cm, minimum height=1cm,text centered, draw=black, fill=red!30]
\tikzstyle{io} = [trapezium, trapezium left angle=70, trapezium right angle=110, minimum width=3cm, minimum height=1cm, text centered, draw=black, fill=blue!30]
\tikzstyle{process} = [rectangle, minimum width=3cm, minimum height=1cm, text centered, draw=black, fill=orange!30]
\tikzstyle{decision} = [diamond, minimum width=3cm, minimum height=1cm, text centered, draw=black, fill=green!30]
\tikzstyle{arrow} = [thick,->,>=stealth]

\def\hlinewd#1{%
\noalign{\ifnum0=`}\fi\hrule \@height #1 %
\futurelet\reserved@a\@xhline} 
\usepackage{makecell}

\tikzstyle{M} = [rectangle, rounded corners, minimum width=2.5cm, minimum height=.5cm,text centered, draw=black, fill=black!15]
\tikzstyle{L} = [rectangle, rounded corners, minimum width=7.5cm, minimum height=.5cm,text centered, draw=black, fill=black!5]

\tikzstyle{Le} = [rectangle, rounded corners, minimum width=7.5cm, minimum height=.5cm,text centered, draw=black, fill=black!30]

\tikzstyle{Lp} = [rectangle, rounded corners, minimum width=5.5cm, minimum height=.5cm,text centered, draw=black, fill=black!30]

\tikzstyle{Ln} = [rectangle, rounded corners, minimum width=7.5cm, minimum height=.5cm,text centered, draw=black, fill=black!0]

\tikzstyle{ML} = [rectangle, rounded corners, minimum width=6cm, minimum height=2cm, text centered,draw=black, fill=black!15]

\tikzstyle{L1} = [rectangle, rounded corners, minimum width=5.25cm, minimum height=.5cm, text centered,draw=black, fill=black!5]

\tikzstyle{Mi} = [rectangle, rounded corners, minimum width=2.5cm, minimum height=.5cm,text centered, draw=black, fill=black!30]

\tikzstyle{Md} = [rectangle, rounded corners, minimum width=1.5cm, minimum height=.5cm,text centered, draw=black, fill=black!5]

\tikzstyle{Mj} = [rectangle, rounded corners, minimum width=4cm, minimum height=.5cm,text centered, draw=black, fill=black!30]

\tikzstyle{s} = [rectangle, rounded corners, minimum width=1cm, minimum height=.5cm,text centered, draw=black, fill=black!15]

\usepackage{xcolor}
\usepackage[normalem]{ulem}

\usepackage{pgfplots}
\pgfplotsset{width=10cm,compat=1.16}

\begin{document}
\maketitle
\begin{abstract}

In the era of social media platforms, identifying the credibility of online content is crucial to combat misinformation. We present the {CREDiBERT} (CREDibility assessment using  Bi-directional Encoder Representations from Transformers), a source credibility assessment model fine-tuned for Reddit submissions focusing on political discourse as the main contribution.
We adopt a semi-supervised training approach for CREDiBERT, leveraging Reddit's community-based structure. By encoding submission content using CREDiBERT and integrating it into a Siamese neural network, we significantly improve the binary classification of submission credibility, achieving a $9 \%$ increase in F1 score compared to existing methods.
Additionally, we introduce a new version of the post-to-post network in Reddit that efficiently encodes user interactions to enhance the binary classification task by nearly $10 \%$  in F1 score. Finally, we employ CREDiBERT to evaluate the susceptibility of subreddits with respect to different topics. 

\end{abstract}

\section{Introduction}

Social media and online news platforms have drastically altered the landscape of news consumption. Individuals now encounter a diverse array of news through social media platforms, compared to the time when people relied on newspapers and television channels. While the digital age offers unprecedented personalization and speed in news delivery, it also presents a significant risk of receiving inaccurate information. Discerning fake news on social media platforms is a formidable challenge, necessitating sophisticated methodologies to distinguish it from authentic reporting.

Recent natural language processing advances have shown great potential in identifying fake news. Through detailed analysis of language patterns and textual features in news content, natural language processing techniques have demonstrated high precision in differentiating fake news from legitimate reports  \cite{raza2022fake}. This breakthrough has catalyzed new research avenues, thus empowering scientists to delve deeper into the characteristics of fake news and devise robust countermeasures.

% \begin{figure}[h]
% 	\centering
% 		\includegraphics[trim={120 100 150 100},clip,width=\linewidth]{Figures/4subs10.png}
%     \caption{The  post-to-post network for  \text{r/Conservative} (dark red), \text{r/democrats}(dark blue), \text{r/Libertarian} (light blue), and \text{r/Republican} (light red) embedded in to 2D plane. }
%     \label{fig:P2P-Netwoek}
% \end{figure}

Despite advances in fake news detection algorithms through content, the reputation of the article's source in providing credible reports is still one of the key components in assessing the trustworthiness of the article. 
As evidenced by Pehlivanoglu et al. (\citeyear{pehlivanoglu2021role}), historical patterns in reporting accuracy often correlate to potential misinformation or propagandistic intents.{ However, tracking the credibility of sources disseminating news via social media is challenging due to the volume and diversity of content creators compared to conventional news sources.}
% However, with the increasing prevalence of news disseminated directly via social media platforms like YouTube and Twitter, tracking the records for credibility analysis has become challenging due to the variety of content created. 
This proliferation of direct-to-consumer news articles necessitates a novel approach to credibility verification. 

This paper diverges from the predominant focus on identifying fake news in the literature. Instead, we concentrate on discerning the \textit{credibility} of news sources shared to combat the spread of misinformation. 
Our objective is to ascertain the credibility of the sources rather than categorically labeling specific news as true or fake. Although there is no direct, infallible correlation between a source's credibility and news veracity, the news source historical track record can offer valuable insights into its general reliability and editorial practices. Such an analysis, especially when conducted over an extended period of time, can reveal patterns and tendencies that indicate the source commitment to accuracy and journalistic integrity.

 Reddit, a prominent social news website, exemplifies the challenges inherent in news source verification. Reddit hosts various \textit{subreddits}, each focused on a specific topic. Users frequently post submissions in political subreddits, including links to news articles for discussion. Reddit's distinctive platform, where users can anonymously post content in various topic-specific communities, provides a conducive environment for the dissemination of fake news. The substantial proliferation of misinformation about COVID-19 in Reddit and QAnon incidents underscores the importance of mitigating misinformation across Reddit before it spreads to other platforms. The expansion of news sources to social media platforms such as Twitter and YouTube made verifying the credibility of the submission's sources often challenging. While some submissions link to verifiable sources, such as major news websites, many do not. In fact, more than $60 \%$  of submissions posted in major political subreddits do \textit{not} have a verifiable source.

\begin{table*}[t]
\centering
{\renewcommand{\arraystretch}{1.3}
\begin{tabularx}{\linewidth}{Xccc} 
 \Xhline{3\arrayrulewidth}
 \textbf{Submission content} & \textbf{Subreddit} & \textbf{Credibility Score} & \textbf{Overall Score}  \\ 
 % \hline
 \hline
 \shortstack{\\Poll: Majority of Republicans would support  Trump in 2024 \\ \,} & \shortstack{r/politics\\ \,} & \shortstack{0.86\\\,} & \shortstack{78\\ \,}\\
 % \hline
 \shortstack{\\ For everyone who thinks he is going away: Majority 
 of  \\Republicans would support Trump in 2024}	 & \shortstack{r/Republican\\\,} & \shortstack{0.83\\\,} & \shortstack{85\\\,}\\
 % \hline
 \shortstack{\\Trump remains the most popular Republican in the country \\ and the leading candidate  for the 2024 GOP nomination}		 & \shortstack{r/Conservative\\\,} & \shortstack{0.38\\\,} & \shortstack{963\\\,}\\
 \Xhline{3\arrayrulewidth}
\end{tabularx}}
\caption{Example of submissions referencing the same event with various representations in different subreddits}\label{Table:simnews}
\end{table*}

This paper presents {CREDiBERT}, a pre-trained language model designed to evaluate the credibility discrepancies between Reddit submissions. Recognizing that identical news stories are often reported with distinct biases across various communities, we leverage the community-based structure of Reddit to generate a comprehensive dataset. This dataset comprises over 1.3 million pairs of news submissions, each pair referencing the same event but sourced from different outlets. These pairs were extracted from five major political subreddits, thus providing a diverse basis for  {CREDiBERT} to learn and effectively gauge the credibility of similar news narratives. The semi-supervised training process equips  {CREDiBERT} to recognize and assess the nuanced differences in news representation, aiding in identifying credible sources.

As the second contribution, When comparable submissions exist, we assess news sources credibility using a Siamese-like neural network architecture \cite{koch2015siamese}. We encode the submission contents by CREDiBERT. We then compare the submission in question with the anchor submission using a Siamese network. In the submission credibility assessment task, the presented framework outperforms BERT-based sentence classifiers by over $9 \%$  in the F1 score.

As the last contribution, we address the problem of credibility assessment on the Reddit platform when comparable submissions are inaccessible or scarce.  We introduce a  weighted post-to-post network that efficiently represents the social interactions among Reddit users. Of note, this network does \textit{not} rely on user profiling, which often raises privacy and security concerns. We then employ a Graph Convolutional Network architecture to infer the credibility of submission sources based on the surrounding social context. This approach ensures a balance between insightful analysis and user privacy, thus paving the way for more ethical credibility assessment in social media.

This paper is structured as follows: It starts with a review of related works. Next, we describe our dataset, which comprises approximately $1.2$ million Reddit submissions and over $12$ million pairs of submissions. In the first half of the Methods section, we present the CREDiBERT. The second half introduces a new post-2-post network, which incorporates user reactions from comments to enhance the results further. Following this, we compare our framework with other baselines in the Results section and discuss its applications and limitations in the Discussion section. The paper concludes by highlighting our main contributions.

% The rest of the paper is organized as follows. First, we present related works within this field. We then detail the dataset we created, encompassing nearly $1.2$ million Reddit submission texts and over $12$ million pairs of submissions in the next chapter. In the first half of the Methods section, we discuss the  {CREDiBERT} architecture and Siamese network, focusing on predicting the credibility of news sources. In the second half, we unveil a novel post-2-post network, enhancing CREDiBERT's performance by incorporating user reactions from comments to the submission. Then, we cross-value the proposed framework with different baselines in the Results section. We discuss the applications and limitations of the proposed framework in the Discussion section. Finally, we conclude the paper by summarizing our main contribution.

\section{Related Work}

Over the last decade, the issue of fake news detection has garnered substantial interest, particularly highlighted by its implications during major events such as the 2016 U.S. presidential election. Initial strides in the domain were made by Castillo et al. (\citeyear{castillo2013predicting}), who focus on assessing the credibility of Twitter content during crises, thus laying the groundwork for understanding digital misinformation. The momentum was significantly amplified post-2016 by attracting a wide array of research, including social impact assessments by Allcott and Gentzkow  (\citeyear{allcott2017social}) and broader misinformation detection techniques by Shu et al. (\citeyear{shu2017fake}). Chan and Donovan (\citeyear{chan2017debunking}) further extended this realm into strategies for countering misinformation. A persistent challenge noted across studies, particularly by Torabi Asr and Taboada  (\citeyear{torabi2019big}), is the scarcity of high-quality labeled data, which impedes the development of robust detection mechanisms.

% The fake news detection problem has been the subject of interest for over a decade. Castillo, \cite{castillo2013predicting} pioneered work in assessing the credibility of Twitter content during crisis events. The 2016 presidential election in the United States created significant momentum and attracted various researchers to this topic . The problem can be investigated in the broader context of detecting misinformation \cite{shu2017fake} and countering misinformation . One of the main barriers to misinformation detection is the scarcity of high-quality labeled data \cite{torabi2019big}. 

\begin{figure*}
	\centering
        \includegraphics[width=\textwidth]{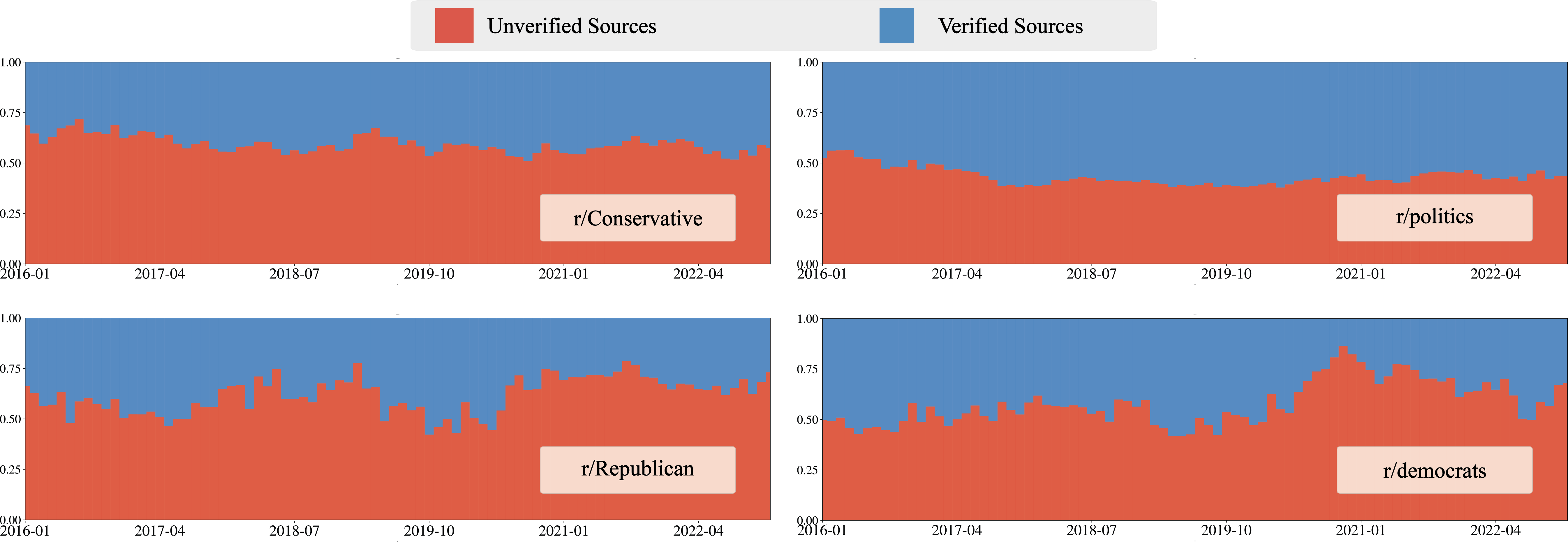}
        \caption{Monthly portion of verified and unverified sources in Reddit. Unverified sources are predominant across subreddits.}
        \label{fig:verification}
\end{figure*}

\subsection{Automated Fact Checking}

Automated fact-checking has significantly evolved with the advancement of artificial intelligence, increasingly relying on the intricate features of news content and social context to discern authenticity. Notably, the introduction of Bidirectional Encoder Representations from Transformers (BERT) marked a substantial advancement in natural language processing, enhancing the capability to process and understand news content. Devlin et al. (\citeyear{devlin2018bert}) spearheaded this development, which was later built upon by Jwa et al. (\citeyear{jwa2019exbake}), who proposed exBAKE, a BERT-based architecture specifically tailored for fake news identification. This model notably improved detection efficacy, as measured by the F1 score. However, the journey of refinement continued as Przybyła (\citeyear{przybyla2020capturing}) critically examined the text style role in automated fact-checking, revealing that while BERT-based methods are potent, they tend to overfit and might underperform in style capturing tasks of large texts compared to  Stylometry and BiLSTM. Addressing some of these concerns, Raza and Ding (\citeyear{raza2022fake}) proposed an enhanced transformer model that leverages both content and social media traces, pushing the boundary further in automated fact-checking. This ongoing evolution, including the rise of pre-trained large language models, continually presents new opportunities and challenges in the field, as evidenced by the work of Chen and Shu (\citeyear{chen2023combating}) and Leite et al. (\citeyear{leite2023detecting}), who explore these cutting-edge developments in combating misinformation.

\subsection{Source Credibility Assessment}

The quest to understand and counteract misinformation has led researchers to examine the role of source credibility closely. Pehlivanoglu et al. (\citeyear{pehlivanoglu2021role}) suggest that historical patterns in a source reporting accuracy can indicate misinformation or propagandistic content, setting a foundation for further studies. Spezzano et al. (\citeyear{spezzano2021s}) explore how users recognize and react to news from low-credibility sources on social media, while Mosleh and Rand (\citeyear{mosleh2022measuring}) shift focus towards measuring user exposure to misinformation rather than identifying false news directly. 
Measuring user exposure to misinformation necessitates a robust method to assess media quality, a controversial and challenging task. 
Bachmann et al. (\citeyear{bachmann2022defining}) contribute to this by proposing a quantitative measure for news media source quality.
Further technological advancements are employed by Chiang et al. (\citeyear{chiang2022investigating}), who apply BERT and artificial neural networks to automate the identification of source credibility. Lastly, Chipidza et al. (\citeyear{chipidza2022ideological}) provide an interesting angle by investigating the interplay between the ideological stance and perceived source credibility, particularly in the context of COVID-19 discussions on Reddit. 
These research studies illuminate various facets of source credibility, highlighting its multifaceted nature and the diverse methodologies employed to understand and evaluate it.

\subsection{Misinformation on Reddit} 

Reddit's unique combination of anonymity and community-driven content, exemplified by incidents like QAnon, fosters an environment conducive to spreading fake news. Understanding the mechanics of this spread is crucial for effective detection and mitigation. Glenski et al. (\citeyear{glenski2018identifying}) shed light on this issue by highlighting how user engagement metrics and subreddit-specific norms significantly influence the perceived credibility of posts on Reddit. Their findings underscore the importance of contextual and community factors in content credibility. Building upon the need for robust data in studying these dynamics, Sakketou et al. (\citeyear{sakketou2022factoid}) introduced the FACTOID dataset, a comprehensive collection of Reddit submissions with annotated credibility and bias information. This dataset facilitates a deeper analysis of how misinformation spreaders operate within the Reddit ecosystem. Bond and Garret's (\citeyear{bond2023engagement}) study paved the way for more nuanced research into users' temporal and contextual interactions with both true and false news submissions on the platform, contributing to the broader understanding of misinformation propagation in online communities.

% The combination of Reddit's anonymity and community structure fosters an environment conducive to spreading fake news such as the QAnan incident. Such incidents underline the importance of misinformation detection across Reddit.
% \cite{glenski2018identifying} highlighted the role of user engagement metrics in predicting the credibility of Reddit posts. Their work emphasized the influence of subreddit-specific norms and user interactions in shaping content credibility. 
% \cite{sakketou2022factoid} introduced the FACTOID, a rich data set for the task of identification of misinformation spreaders. The FACTOID data set includes submissions to various Reddit posts and their corresponding Credit and Bias from the link pointed toward the source. Further studies delve into how users interact with True and fake news submissions across Reddit from temporal and contextual standpoints.

% \paragraph{This paper} stands out from the literature in two aspects. First, we leverage the unique community structure of Reddit to create a data set of pairs of submissions that point toward the same news. Second,  the  {CREDiBERT} model which is trained semi-supervised, utilizes Siamese network structure and manages to overcome the overfitting challenges mentioned above for BERT-Bace models. 

\textbf{This paper} distinguishes itself from the existing body of literature by introducing two significant innovations: First we leverage Reddit's intricate community structure to automatically curate a novel dataset comprising pairs of submissions referencing identical events, providing a unique vantage point for understanding source credibility. Secondly, we introduce the 'CREDiBERT' model, a semi-supervised approach utilizing a Siamese network architecture specifically designed to mitigate the overfitting issues prevalent in standard BERT-based models for this task  \cite{przybyla2020capturing}. This adaptation allows for a more robust and generalizable source credibility assessment, addressing a critical challenge in the field.

\section{Data}

We compile a comprehensive dataset from five major political subreddits: \text{r/politics}, \text{r/Conservative}, \text{r/Libertarian}, \text{r/Republican}, and \text{r/democrats}. Our dataset, covering the period from January 2016 to December 2022, includes detailed information such as submission text, author IDs, source domains, submission times,  associated subreddits, overall submission scores, and comment counts for each submission\footnote{All Reddit data collected from \url{https://pushshift.io}}. We then collect all available comments from these submissions. This extensive dataset forms the basis for our research into detecting submission source credibility and provides a rich resource for training  {CREDiBERT}. We select these specific subreddits based on their activity volume to ensure a diverse range of political discussions and viewpoints. In collecting this data, we adhere to ethical guidelines for research involving online communities, ensuring the anonymity and privacy of the users' data.

We utilize a dataset from the Ad Fontes Media website\footnote{ Details available in \url{https://adfontesmedia.com/}}, which provides Bias and Credibility scores for $223$ major news sources. To enhance the reliability of this data, we cross-validate these scores with additional datasets from FACTOID date set  \cite{sakketou2022factoid} and Media-Bias-Fact-Check website\footnote{
Accessible here \url{https://mediabiasfactcheck.com/}
} for the common sources. We consider a news source as \textit{verified} if it appears in our list of $223$ sources with available credibility and bias scores. The credibility scores range from 0 to 1, with 0 being the least credible and 1 being the most credible source.   We label a submission as noncredible if its source credibility score is less than $0.6$. While we consider established news outlets like the New York Times and Fox News to be credible, we acknowledge the potential for bias and inaccuracies in any source, particularly when it comes to political content.

\begin{table}[h]
    \centering
    {\renewcommand{\arraystretch}{1.3}
    \begin{tabularx}{230pt}{Xccc}
        \Xhline{3\arrayrulewidth}

        % \vspace{2mm}
        \textbf{Subreddit} \hspace{10mm}& \textbf{\# Submission} & \textbf{Verified} & \textbf{Unverified} \\
        
        \hline
        \text{ r/Politics} & $2,485,983$ & 48\%& 52\% \\
        \text{r/Conservative} & $788,293$ &  28\%& 72\%  \\
        \text{r/Libertarian} & $29,6402$ &  84\%& 16\% \\
        \text{r/democrats} & $147,597$ &  23\%& 77\% \\
        \text{r/Republican} & $142,570$ & 28\%& 72\% \\
        % dailywire.com & $22,825$ & \ding{51}\\
        % nypost.com & $13,117$  & \ding{51} \\
        % redstate.com & 11292 & YES \\
        % townhall.com & 11143  & YES \\
        \Xhline{3\arrayrulewidth}
    \end{tabularx}}
    \caption{Overall number of submissions in five major political subreddits, with the ratio of verified and unverified sources during $7$ years.}\label{Table:subreddits_ver_ratio}
\end{table}

 Our analysis includes statistics on the top $5$ predominant news sources for submissions in \text{r/Conservative}, drawn from a pool of $18,015$ different sources, as presented in Table \ref{Table:conservative_source}. We also examine the distribution of verified and unverified sources across the studied subreddits, shown in Table \ref{Table:subreddits_ver_ratio} and Figure \ref{fig:verification}. This analysis reveals that the ratio of verified to unverified sources remains relatively stable over our study period across all subreddits. For the purposes of this research, we focus only on submissions with comments and those from verified sources, as these provide the most reliable data for assessing news source credibility.

After thorough data cleaning, the dataset comprises a total of $1,312,853$ submissions. We organize comments based on their reply level: first-tier comments are direct responses to submissions, second-tier comments are replies to first-tier comments, and so on. We limit our analysis to comments up to the fourth tier, as we observed that comments beyond this level often diverge significantly from the original submission content. To maintain data quality, we exclude comments when the author's account has been deleted or removed, or if the comment text is unavailable. We omit comments from \text{r/politics} due to their disproportionately large volume, which could skew the analysis and present unmanageable computational burdens. This exclusion resulted in a final count of $3,879,140$ comments from nearly $81,000$ authors. We train the CREDiBERT on submissions and comments dated before $2022$ to ensure robust model testing with sufficient unseen data. The experiments utilize data collected from the $2022$, which includes $59,637$ submissions. This temporal split in training and testing data helps to evaluate the model performance on recent, previously unseen data, providing a more reliable assessment of its predictive capabilities.

\begin{table}[t]
    \centering
    {\renewcommand{\arraystretch}{1.3}
    \begin{tabularx}{230pt}{Xcc}
        \Xhline{3\arrayrulewidth}        
        % \vspace{2mm}
        \textbf{Source Website} \hspace{10mm}& \textbf{\# Submission} & \textbf{Verifiable}\\
        
        \hline
        self.Conservative & $108,055$ & \ding{55} \\
        i.redd.it & $82,875$ & \ding{55} \\
        youtube.com & $39,649$ & \ding{55} \\
        foxnews.com & $28,763$ & \ding{51}\\
        breitbart.com & $25,666$ & \ding{51} \\
        % dailywire.com & $22,825$ & \ding{51}\\
        % nypost.com & $13,117$  & \ding{51} \\
        % redstate.com & 11292 & YES \\
        % townhall.com & 11143  & YES \\
        \Xhline{3\arrayrulewidth}
    \end{tabularx}}
    \caption{Predominant referenced sources for submissions in \text{r/Conservative} from January $2016$ to December $2022$.}\label{Table:conservative_source}
\end{table}

\section{Method}

This section outlines our method for assessing the credibility of submission sources on Reddit.
It is important to note that when we discuss a submission credibility, we specifically refer to its source credibility. To pair submissions that reference the same events but originate from different sources, we employ Sentence Transformers \cite{reimers2019sentence}. This pairing is critical as it allows for a comparative analysis of how different communities report identical events. We then use this paired dataset to train {CREDiBERT}, a sentence embedding model trained based on the Siamese transformer architecture. This training approach is specifically designed to evaluate the credibility of submission by closely aligning submission texts with minimal credibility discrepancies and distancing those with significant discrepancies. This method establishes a robust and comparative framework to understand and quantify submission credibility variations on Reddit.

\subsection{Submission Pairing}\label{subc.pair}

Moderators play a pivotal role in shaping the content of each subreddit, thus ensuring that submissions align with the subreddit intended narrative and bias. 
This moderation influences the diversity of news sources and perspectives presented within their domains. 
% To understand how this moderation affects the credibility of news sources, we focus on pairing submissions from different subreddits. 
We utilize sentence transformers (S-BERT) introduced by Reimers and Gurevych \citeyear{reimers2019sentence} for this task, which allows a comparison of how similar news events are reported differently across various communities.
As shown in Table \ref{Table:simnews}, we can observe significant discrepancies in news credibility scores when comparing similar news shared across different subreddits. This approach enables us to pair sentences and assess their credibility score discrepancies automatically.

While the BERT-based model excels at creating context-aware word embeddings through training with a masked language model and next-sentence prediction, it falls short in efficiently embedding entire sentences in the latent space. To overcome this, Reimers et al. \citeyear{reimers2019sentence} introduced a modification known as sentence transformers, which uses a Siamese network architecture to enrich sentence-level embedding. The resultant S-BERT model outperforms traditional BERT-based models in embedding sentences. In this study, we employ the pre-trained ``all-distilroberta-v1" model from the pre-trained S-BERT library in the huggingface website, chosen for its robustness in capturing sentence nuances. This tool is pivotal in our method for identifying and pairing submissions that are thematically similar yet sourced differently, thus contributing significantly to our analysis of credibility assessment across Reddit submissions.

 We define the complete set of collected submissions as $\mathcal{S}:=\{s_1,s_2,\cdots ,s_n\}$, where $n$ represents the total number of submissions. For each submission $s_i$, we denote its text, credit score, and posting time as $p_i$, $c_i$, and $t_i$, respectively. To determine the similarity between two news submissions, we encode the text $p_i$ through a sentence transformer, denoted by $\bf T$. 
 This model maps the content of submission $s_i$ to the embedding $e_i$, by $e_i = {\bf T}(p_i)$, where $e_i$ is a  vector of real numbers with ${768}$ dimensions. We utilize the bi-encoder technique to identify similar submissions, which assesses the similarity between two submissions, $s_i$ and $s_j$, by the cosine similarity of their corresponding embeddings. We calculate the similarity by
\begin{equation}
    \cos(e_i,e_j)= \frac{\langle e_i , e_j\rangle}{\|e_i\|\|e_j\|},
\end{equation}
 where $\|\cdot \|$ and $\langle \cdot \rangle$ denote the Euclidean norm and the inner product, respectively. This method allows us to quantitatively compare submissions based on their textual content, hence enabling a more precise identification of similar news stories across different subreddits.

We opt for bi-encoders over cross-encoders to efficiently compare the vast number of submission pairs, primarily due to their faster processing capabilities.  Reimers and Gurevych \cite{reimers2019sentence} note that comparing $10,000$ pair of sentences using cross-encoders would take about $65$ hours while generating embeddings and computing cosine similarities would take less than $6$ second. Given that our dataset encompasses over one million submissions, this choice significantly reduces the computational burden.

Recognizing that submissions referencing the same event are likely posted in close temporal proximity, we implement a time constraint, denoted as $\Delta$, to refine our search. This constraint means we only consider pairing submissions if their posting times, represented by $t_{ij}=|t_i-t_j|$, are within $\Delta$. 

Two submissions are similar if they meet the temporal criteria and their similarity score is more than the similarity threshold $\bar{e}$.  For each qualified pair of similar submission $(s_i,s_j)$, we then calculate the credit score difference $c_{ij}=|c_i-c_j|$, which is the absolute difference in their credit scores. This approach streamlines our process, ensuring we focus on our analysis's most relevant and temporally aligned submission pairs.

In our methodology, we define the pool of all pairs of similar submissions, $\mathcal{P}$, by
\begin{equation}\label{def.setP}
\mathcal{P}:=\{(s_i,s_j,c_{ij})|~ \bar e < \cos(e_i,e_j) , ~|t_i-t_j|\leq \Delta\}.
\end{equation}
This approach effectively expands the dataset from $1.2$ million individual submissions to a comprehensive set of $12$ million uniquely paired submissions. This significant increase in data pairs permits us to train a robust language processing model, enhancing the validity of the model in examining the credibility of submissions across Reddit.

\begin{figure}
    \centering
\begin{tikzpicture}[node distance=2cm]

\node (credibert1) [M] { \footnotesize CREDiBERT};

\node (submission1) [Mi, below of=credibert1,yshift=1cm] {\footnotesize Submission};

\node (event) [Lp, below of=submission1, yshift=1cm,xshift=3.5cm] {\footnotesize  Submission Pool $\mathcal{P}$};

\node (submissionL) [Le, below of=event,yshift=1cm,xshift=-1cm] {\footnotesize  Unlabeld Submission};

\node (submission2) [Mj, right of=submission1, xshift=2.25cm] {\footnotesize  Anchor Submission};

\node (credibert2) [M, right of=credibert1,xshift=1.5cm] {\footnotesize CREDiBERT};

\node (ei) [fill=white, above of=credibert1,yshift=-1cm] {};
\node (ej) [fill=white,text centered, above of=credibert2,yshift=-1cm] {};

\node (siamese) [ML,above of=ei, yshift=0cm,xshift=1.75cm, text depth = 2 cm, draw,anchor=north] {\footnotesize Siamese Network};

\node (Dense) [L, above of=siamese, yshift=0.cm,,xshift=.75cm] {\footnotesize  Classfication Layer};

\node (Densej) [Md, above of=ej, yshift=-2cm,] {\footnotesize  Dense Layer};

\node (Densei) [Md, above of=ei, yshift=-2cm] {\footnotesize  Dense Layer};
\node (L1)[L1, above of=Densei , yshift=-1.cm,xshift=1.75cm]{\footnotesize L1 Layer};

\node (Label) [Ln,above of =Dense, fill=white,yshift=-1cm] {\footnotesize Submission Credibility Label};

% \node(4)[ML,text centered,text width = 0.25\textwidth, text depth = 2 cm,
%     draw,above of=Label,anchor=north]{title is very long and long};
% \node (score) [s, right of=credibert2,xshift=0.25cm] {Score};
% \node (cosine) [M, above of=credibert1, xshift=2cm,yshift=-0.5cm] {Similarity};
\draw [arrow] (submission1) ->(credibert1);
% \draw [arrow] ([xshift=-2.5cm]event.north) -- (submission1.south);
% \draw [arrow] (ei.north) -- (Densei.south) ;
% \draw [arrow] (ej.north) -- (Densej.south) ;
\draw [arrow] ([xshift=-.75cm]submission2.north) -- (credibert2);
\draw [arrow] ([xshift=1.25cm]submission2.north) -- ([xshift=3cm]Dense.south) node [below=107pt, fill=white] {\scriptsize Credibility Score};
\draw [arrow] (Densei.north) -- ([xshift=-1.75cm]L1.south);
\draw [arrow] (Densej.north) -- ([xshift=1.75cm]L1.south);

\draw [arrow] (siamese.north) -- ([xshift=-.75cm]Dense.south);
% \draw [arrow] ([xshift=1.5cm]submission2.north) -- (score);

\draw [arrow] (credibert1) -- (Densei);
\draw [arrow] (credibert2) -- (Densej);
% \draw [arrow] (credibert2) -- (cosine);
% \draw  [->] (submission2) -- (score);
\draw[arrow] ([xshift=.75cm]event.north) -- ([xshift=0cm]submission2.south);

\draw [arrow] ([xshift=-2.5cm]submissionL.north) -- (submission1.south);

\draw[arrow] ([xshift=1.75cm]submissionL.north) -- ([xshift=.75cm]event.south);

\draw [arrow] (Dense) -- (Label);
Label
\end{tikzpicture}
    \caption{Fine-grained Siamese network architecture for credit assessment when an anchor submission is accessible.}
    \label{fig:Siamse}
\end{figure}
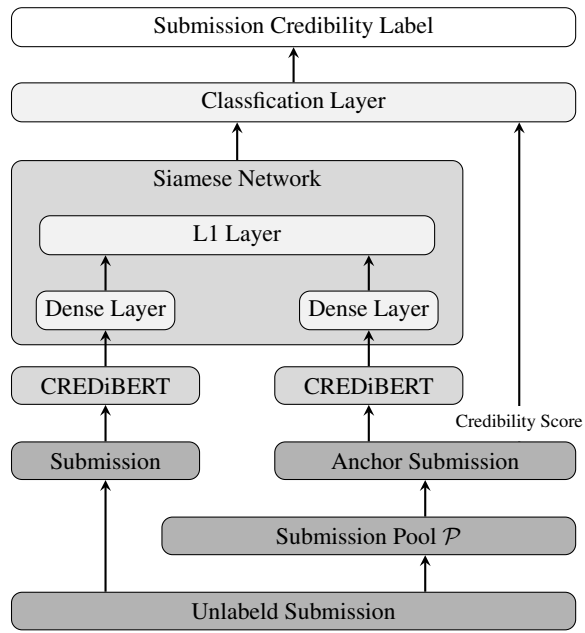

% \subsection{CREDiBERT}
\subsection{CREDiBERT}

To effectively gauge the credibility of the submissions on Reddit, we introduce {CREDiBERT}, a specialized transformer-based model designed for this task. Unlike conventional approaches that classify submissions as simply credible or not, {CREDiBERT} is uniquely focused on predicting the credit score discrepancy between two submissions referencing the same event. This nuanced approach allows for a more refined analysis of credibility. 
The underlying objective is to align the embeddings from {CREDiBERT} such that cosine similarity between pairs with minor credibility score discrepancies have values close to 1, indicating similarity, while those with significant discrepancies are marked as dissimilar. To ensure a balanced training set, we meticulously selected $1,383,385$ samples from the pool of $12,171,894$ pairs. This selection was guided by the goal of achieving representative and diverse training data, thereby enhancing the reliability and generalizability of the {CREDiBERT} model's assessments.

{To train the {CREDiBERT} model, we adopt a Siamese architecture similar to S-BERT. This choice is motivated by the architecture effectiveness in processing and comparing text embeddings when pairs of labeled sentences are available.} The training is guided by the following objective function, 
\begin{equation}
\sum_{(i,j)\in \mathcal{P}_\tau} | \cos({\bf C}(p_i),{\bf C}(p_j)) - \big(1-c_{ij}\big) |^2,
\end{equation}
where ${\bf C}(\cdot)$ denotes the embedding generated by  {CREDiBERT}, and $\mathcal{P}_\tau$ represent the set of training pairs. The objective function aims to minimize the discrepancy between the cosine similarity of the embeddings and the actual credit score differences. 
We trained the CREDiBERT model using the 'all-distilroberta-v1' S-BERT model as the starting point with a dataset of $1,312,853$ pairs of submissions. The entire training process took nearly 8 hours on a single NVIDIA A100 GPU.

\paragraph{Siamese Network }  In cases where a news event significantly impacts various communities and recurs across them, we can identify an \textit{anchor} submission. This submission is selected as the most representative or similar among others in the set $\mathcal{P}$, calculated by
\begin{equation}
a_i = \argmax_{j\in \mathcal{P}_i} \big(\cos(e_i,e_j)\big),
\end{equation}
where $\mathcal{P}_i$ comprises submissions paired with submission $i$. The anchor submission, originating from a known source, provides a reliable credit score for comparison. We then use a secondary Siamese network to predict the credit score discrepancy between each pair, followed by a dense neural network layer to classify the news as credible or not. Figure \ref{fig:Siamse} depicts the framework for the fine-grained Siamese Network, designed for credit assessment tasks. This approach ensures that our model accurately captures the nuances of credibility across various submissions on Reddit.  The Dense layers are a combination of two layers with $32$ and $16$ hidden nodes each. The classification layer comprises a single rely layer with $8$ hidden state followed by a softmax layer.

\begin{table*}[t]
    \centering
    {\renewcommand{\arraystretch}{1.3}
    \begin{tabularx}{505pt}{XXccccc}
        \Xhline{3\arrayrulewidth}
        {Model Type}& \multicolumn{1}{c}{\textbf{Model}} & \multicolumn{1}{c}{\textbf{Anchor}}&  \multicolumn{3}{c}{\textbf{Accuracy}}& \multicolumn{1}{c}{\textbf{F1}}  \\
        & & & Non-Credible&Credible& Overall &\\
        \hline
        \multirow{2}{*}{\textit{Random}}&Random  & NO& $0.496$& $0.512$ &$0.499$ & $0.459$ \\
        &Majority   & NO& $0.000$ & $1.000$& $0.771$ & $0.435$\\
        \hline
        \multirow{2}{*}{\textit{Word Embedding}}&word2vec-google-news-300& NO & $0.551$ &$0.765$ &$0.707$ & $ \bf 0.650$\\
        &fasttext-wiki-news-300 & NO& $0.531$& $0.765$&$0.700$ & $0.6416$\\
        \hline
        \multirow{3}{*}{\textit{BERT-based Text Classifier}}& BERT-base-uncased   & NO&$0.579$ & $0.878$& $0.775$ & $ \bf 0.738$\\
        &DistilBERT-base-uncased   &NO& $0.566$&  $0.880$& $0.777$ & $ 0.734$\\
        &RoBERTa-large   &NO&$0.423$ & $0.938$& $0.763$ & $ 0.694$\\
        
        \hline
        \multirow{3}{*}{\textit{Sentence Transformer}}&S-BERT: all-distilroberta-v1 & NO& $0.422$&$0.912$ &$0.775$ &  $0.683$ \\
        &S-BERT: all-MiniLM-L12-v2  & NO& $0.694$& $0.757$&$0.713$ &  $0.{669}$ \\
         &{CREDiBERT}  (\textit{Ours})  & NO& $0.680$& $0.922$ &$0.856$ &  ${ \bf 0.814}$  \\
         
        \hline
         \multirow{2}{*}{\textit{Siamese Network}} &S-BERT: all-distilroberta-v1  & YES& $0.460$ & $0.920$&$ 0.792$ & $ 0.708$  \\
         &{CREDiBERT}   (\textit{Ours})  & YES& $0.713$ & $0.914$&$0.861$ & $ \bf 0.822$\\
        \Xhline{3\arrayrulewidth}
    \end{tabularx}}
    \caption{CREDiBERT surpasses other text classification models in binary credibility classification of submissions, as evidenced by cross-validation results. We train and validate all models using $47,710$ and $5,963$ samples, and reporter results on $5,963$ test samples. We show the best F1 performance in each class in bold. 
    }\label{Table:Results-sim}
\end{table*}

\subsection{Post-to-Post Network}

 While major events receive broad coverage across communities, many topics are confined to a few related subreddits, leading to a scarcity of submissions that meet the established similarity criteria. We develop a post-to-post network based on user interaction patterns to overcome this challenge and further enhance the results. The underlying premise is that if users exhibit similar reactions to two posts(submissions), these posts imply some similarity. Ideally, one could encode these reactions alongside text embeddings and analyze them through a model to discern subtle similarities. However, this approach poses a significant challenge due to the sheer volume of user engagement, with tens of thousands of active authors commenting on submissions. To address this issue, we explore an alternative method to distill meaningful patterns from user interactions by focusing on key indicators that reflect the essence of users reactions without getting overwhelmed by the data volume. 
 % This approach allows us to extend our analysis to a wider range of topics, capturing the nuances of user engagement and its implications for post similarity within the Reddit ecosystem.

% While major events would be covered across all communities, many subjects are only covered by a few related communities. Thus, there are no submissions that meet the similarity conditions. To address this issue, we create a post-to-post network based on the interaction of users. We created the network because if users react similarly to two posts, these two submissions share similarities. One can encode users' reactions across the text embedding and pass them all through a model to learn the results. However, tens of thousands of users actively comment, making it extremely challenging, if impossible.  

We present a new version of the post-to-post network introduced by Hurtado et al. (\citeyear{hurtado2019bot}). The post-to-post network is based on the premise that submissions with similar commenting patterns, particularly regarding their authors, are likely to be more closely related. While the original network considers submissions connected if they share at least one commenter, Bond and Garret  (\citeyear{bond2023engagement})  suggest that authors may exhibit different reactions depending on the news authenticity. We employ S-BERT to encode the comment content in order to estimate the users reactions toward the submission. For instance, if a comment $q_k^i$ on submission $s_i$ responds to a parent comment $q_{l}^i$, we measure the similarity between these comments by the cosine similarity of their embeddings, $\cos({\bf T}(q_{l}^i),{\bf T}(q_{k}^i))$, recalling ${\bf T}(\cdot)$ represent the S-BERT encoding model. We can then calculate the  submission-similarity for comment $q_k^i$ toward the submission $s_i$ by 
\begin{equation}
\alpha_{k}^i=\alpha_{l}^i \times \cos({\bf T}(q_{l}^i),{\bf T}(q_{k}^i)),
\end{equation}
where $q_{l}^i$ is the parent comment for $q_{k}^i$.
This method enriches the analysis by considering the depth of user engagement and its impact on the perceived similarity between various submissions.

We initiate the analysis with first-tier comments, assigning a submission-similarity score based on the cosine similarity between the submission and the comment, calculated as $\cos({\bf T}(p_i),{\bf T}(q^i))$ for some first tier comment $q^i$, posted on submission $s_i$. This process is then iteratively applied to second-tier comments and subsequent tiers, allowing us to determine the post-similarity score for all comments across different levels.
To encapsulate the spectrum of user reactions for each post, we construct a vector, $r_i$, which aggregates the reactions of all users within the set $\mathcal{A}$. For an author $a$ who has commented on submission $s_i$, the reaction is quantified in the  $a$th element of the vector $r_i$ by 
\begin{equation}
    \{r_i\}_a =\frac{1}{|\mathcal{C}_a^i|} \sum_{k \in \mathcal{C}_a^i}\alpha_{k}^i,
\end{equation}
where $\mathcal{C}_a^i$ is the set of all comments author $a$ posted in reply to submission $s_i$.
We denote the cardinality of the set $\mathcal{C}_a^i$, the number of distinct elements in the set,  by $|\mathcal{C}_a^i|$. In cases where an author has not commented on submission  $s_i$, we set $\{r_i\}_a =0$. Given the large number of authors, represented by $|\mathcal{A}|$, we navigate the challenge of combining the embedding vector and the reaction vector for each submission by introducing a weighted version of the post-to-post network.

To better capture the intricate interplay between user interactions and posts, we define the post-to-post graph $\mathcal{G} := (\mathcal{V},\mathcal{E},\mathcal{W})$. In this graph, $\mathcal{V}$ denotes the nodes (submissions), with $\mathcal{E}$ and $\mathcal{W}$ representing the edges and weights, respectively. We consider the set of users who commented on post $i$ as $\mathcal{A}_i$. An edge is established between submissions $i$ and $j$ if they share more than $m$ common users, leading to the edge set $\mathcal{E}$. Thus $\mathcal{E}$ is  defined as
\begin{equation}
    \mathcal{E} :=\{ (s_i,s_j)~|~~ \lvert\mathcal{A}_i \cap \mathcal{A}_j\rvert > m\}.
\end{equation}
This structure allows us to analyze the relationships and similarities between submissions based on user engagement patterns.
While the original post-to-post network captures the basic structure and offers insights into the connections between posts, it does not differentiate between positive and negative user reactions to submissions. We enhance the network by assigning weights to its edges to address this. Specifically, for an edge $(i,j) \in \mathcal{E}$, we calculate the weight $w_{ij}$ by
 \begin{equation}
     w_{ij} = \langle r_i,r_j\rangle,
 \end{equation}
where $\langle \, \cdot \,, \, \cdot \, \rangle $ denote the inner product between two vectors. This weight reflects the degree of similarity in user reactions to both submissions. The weights are then stored in $\mathcal{W}$. The innovative approach in assigning these weights incorporates the user responses, thereby enriching the network analytical depth. Importantly, this method avoids profiling individual authors, thus upholding user privacy and safety. For example, if a user exhibits negative and positive reactions to two distinct submissions, our model interprets this as indicating a strong separation between these submissions, while the original network presented by \citeauthor{hurtado2019bot} (\citeyear{hurtado2019bot}) considers them strongly connected. This nuanced analysis allows us to map the network of submissions more accurately, considering the presence of user interactions and their qualitative nature.

\section{Result}

We conduct three distinct experiments to assess the credibility of the submissions using the proposed model, each designed to test different aspects of the {CREDiBERT} and the weighted post-to-post network. In the first experiment, our objective is to assess the credibility of a submission when it is related to at least one other submission covering the same event. We integrate the {CREDiBERT} with the Siamese and feed-forward neural networks to predict credibility labels, focusing on submissions with at least one anchor in the data set. For the second experiment, we relax the existence of anchor assumption to encompass a broader range of submissions. In doing so, we integrate the post-to-post network to analyze and assess submission credibility in a more comprehensive manner. Finally, we demonstrate the capability of the proposed model by performing a case study to detect the susceptibility of the subreddit communities to misinformation with respect to different topics.

 The criteria for two submissions to be similar is to have a cosine similarity of at least $0.6$ and a time difference of no more than $15$ days. In the following sections, we refer to the classification layer, which consists of a feed-forward neural network with two dense layers followed by a softmax layer for binary classification tasks.

\subsection{Credibility Assessment}

  We conduct a comparative analysis of different models when anchor submission is available. We focus on comparing the proposed model, {CREDiBERT}, against the word2vec embedding model \cite{mikolov2013efficient}, standard BERT text classification model \cite{devlin2018bert},  and the S-BERT embedding model \cite{reimers2019sentence}. We emphasize that submission texts are often short and include only one sentence. Thus, methods such as bag-of-words and stylometry, which require medium to large text corpus, are incompatible with this task \cite{przybyla2020capturing}. 
 
 The experiment data set is solely collected from 2022, comprising a total of $59,637$ submissions, divided into training ($80 \%$), validation ($10 \%$), and testing ($10 \%$)  sets. All the models are trained over the identical training data set, and the results are reported over the test data set. This data selection was made to ensure that both the training and test datasets are current and relevant, providing a robust basis for our evaluation. Notably, we train {CREDiBERT} on submission pairs posted before 2022, thus ensuring that our training and test datasets are novel and independent.
 This approach allows us to thoroughly assess the performance and accuracy of {CREDiBERT} in a contemporary context, offering insights into its effectiveness compared to other established models.

\paragraph{Submission Labeling}

We categorize submissions with a credibility score lower than 0.6 as non-credible, while others are deemed credible. This threshold is chosen based on the analysis of the credibility chart and verification of the references. The binary classification is used to determine whether the source of a submission could be considered credible or not. Given the unbalanced nature of our dataset, which is skewed towards credible submissions, we choose the F1 score as the principal metric for comparing model performance rather than relying solely on accuracy.

\begin{figure}[t]
	\centering
		\includegraphics[trim={0 0 0 0},clip,width=1\linewidth]{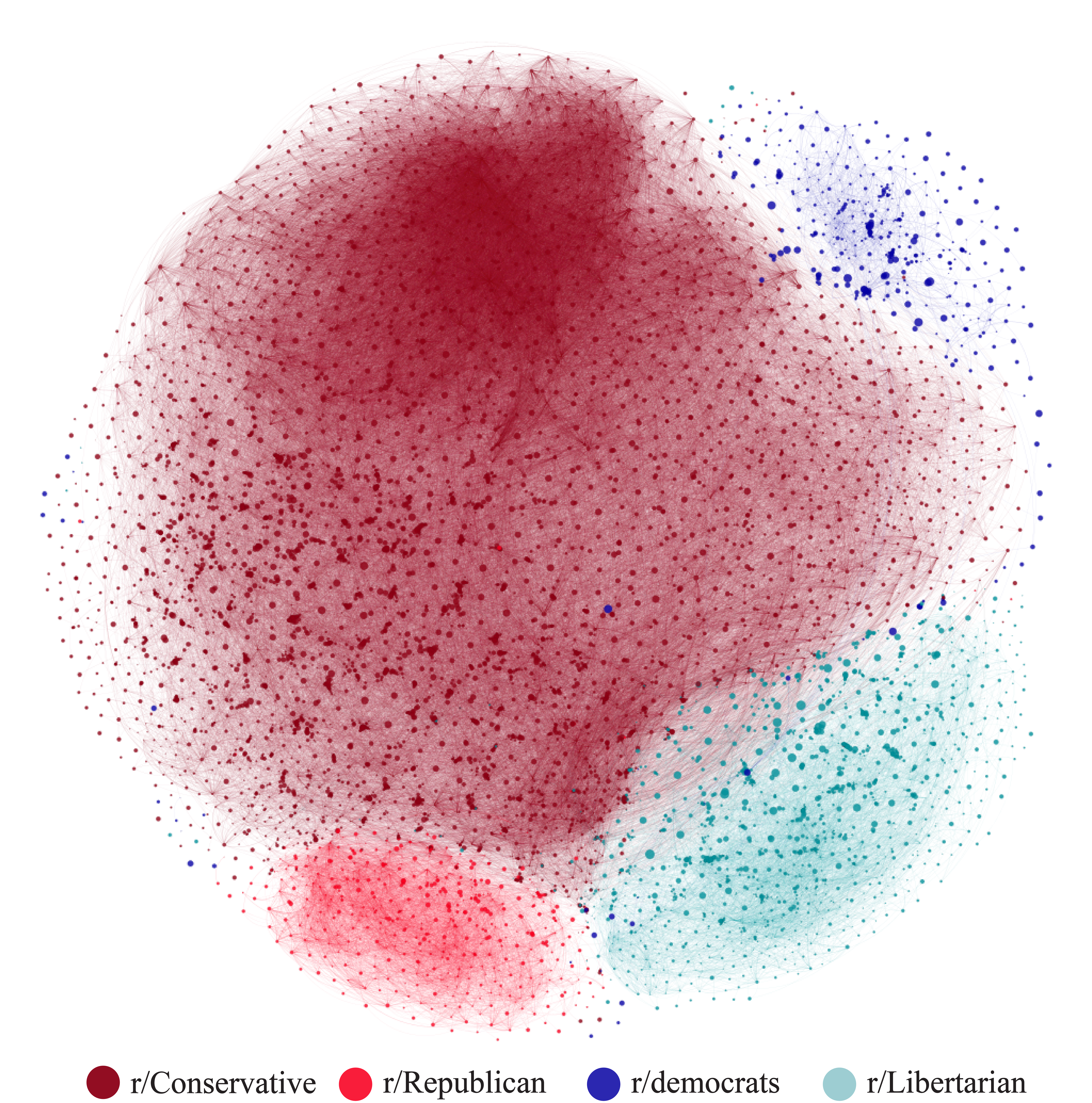}
    \caption{The post-to-post network for $7,466$ submission. The network shows a strong separation between different communities. For brevity, only edges with a weight over $0.3$ are shown.}
    \label{fig:P2P-Netwoek}
\end{figure}

\paragraph{Word2Vec}
Word2Vec was developed as a technique to estimate the vector representation of words efficiently \cite{mikolov2013efficient}. 
We compute the average word embeddings of the submission text and then feed the embedding to a classification layer to assess the credibility of the submissions.

\paragraph{BERT-based Text Classifier} 
  The BERT-based text classifiers fine-tune the underlying transformer for the classification task and utilize the first token as the aggregate sentence representation for classification, allowing them to overperform naive classification of average embeddings pooling generated by BERT-based models.

\paragraph{Sentence Transformers}
The S-BERT model was developed to address the challenges the BERT-based model faces in sentence-level embedding in sentence comparison tasks. We integrate the sentence-level embedding with two different classification models for better comparison. We emphasize that while the CREDiBERT is primarily created for classification tasks,  the model is trained to estimate Credibility score discrepancies; thus, it is considered a sentence transformer model. For the first approach, we integrate sentence-level embeddings with a classification layer for binary classification. Additionally, we employ the Siamese network architecture, depicted in Figure \ref{fig:Siamse} for binary classification.

% Through this comparative analysis, we aimed to assess the effectiveness of {CREDiBERT} in relation to other text classifiers described above to produce meaningful and discriminative embeddings for our specific classification task.

Table \ref{Table:Results-sim} presents the cross-validation results for the binary classification task of assessing submission credibility. In our analysis, employing the Siamese network has enhanced the learning process from sentence embeddings for both S-BERT and {CREDiBERT}. As expected, the BERT-based text classifiers outperformed the pre-trained sentence embedding models in the F1 score. Among the BERT-based classifiers, ``BERT-base-uncased" shows a slight edge over others, specifically in the identification of non-credible sources. As expected, the word2vec model could not perform competently since it does not consider the style and tone of the text and only focuses on the words in the text.

As detailed in Table \ref{Table:Results-sim}, the proposed {CREDiBERT} model marks a significant improvement, outperforming the others in terms of both accuracy and F1 score. The data indicates a near $9 \%$  improvement in  F1 score and $9 \%$  in overall accuracy. The Siamese network and CREDiBERT combination most accurately identify submissions with non-credible sources and outperform all other models in the F1 score.
However, when comparing the efficacy of the Siamese network, its impact is less pronounced with {CREDiBERT} than with S-BERT, suggesting that {CREDiBERT} embeddings already reflect features regarding the credibility and are less affected by contrast with anchor submission.

\begin{table}[t]
    \centering
    {\renewcommand{\arraystretch}{1.3}
    \begin{tabularx}{230pt}{Xcc}
        \Xhline{3\arrayrulewidth}
        % \vspace{2mm}
        \textbf{Model} \hspace{10mm}& \textbf{Acc.} & \textbf{F1}  \\
        \hline
        Random  & $0.505$ & $0.506$ \\
        Majority  & $0.543$ & $0.352$\\
        \hline
        Node2vec+ Dense & $ 0.693$ & $ 0.677$  \\
        \hline
        S-BERT + Dense & $0.708$ &  $0.694$ \\
         {CREDiBERT}  + Dense (\textit{Ours})  & $0.738$ &  $0.738$  \\
        \hline
        S-BERT + GCN  & $ 0.795$ & $ 0.690$  \\
         {CREDiBERT} + GCN  (\textit{Ours})  & $ \bf0.851$ & $\bf 0.796$\\
         \Xhline{3\arrayrulewidth}
    \end{tabularx}}
    \caption{ Cross-validation report for binary classification of $7,468$  submissions over 2017 in four major political subreddits. We train, validate, and test all models with $6,000$,  $743$, and $743$ samples. Incorporating post-2-post network improves results for both S-BERT and CREDiBERT models.}\label{Table:p2p.result}
\end{table}

\subsection{Post-to-Post Network}

In this section, we focus on investigating the effects of incorporating a weighted post-to-post network. We conduct a comparative study of the {CREDiBERT} model against the S-BERT model and node2vec embedding model. Previous experiments have shown {CREDiBERT}’s superior performance in embeddings for the Credibility classification task; therefore, our aim is to investigate how this advantage plays out when augmented with the graph-based structure of the post-to-post network.

We integrate the post-to-post network along with a Graph Convolutional Network (GCN) \cite{kipf2016semi} for the binary classification of submissions. We focus on submissions from the year 2017 across four specific subreddits: \text{r/Conservative}, \text{r/democrats}, \text{r/Libertarian}, and \text{r/Republican}. This selection is driven by the representativeness and diversity of political discourse in these forums. We exclude \text{r/politics} due to its substantial size, which poses significant computational challenges.

Figure \ref{fig:P2P-Netwoek} illustrates the resulting post-to-post network for the $7,468$ submissions encoding more than $120,000$ comments. To select submissions, we first choose the authors with more than 5 comments during $2017$ and then select the posts with at least $5$ comments from the selected authors to ensure meaningful interaction. After cleaning data, we are left with $5,106$  authors and $7,468$  submissions.
In this visualization, the color of each node indicates the subreddit of origin, providing insights into subreddit-specific trends and interactions. Figure \ref{fig:P2P-Netwoek} representation allows us to observe how well a post-to-post network can distinguish between subreddits and reveal the submission connection across different subreddits.

\paragraph{Node2vec}
Node2vec, introduced by Grover and Leskovecis (\citeyear {grover2016node2vec}), is a graph embedding technique. It effectively represents graph nodes as vectors in a latent space, capturing both the structural characteristics and homophilic tendencies of the graph. We pass the graph embeddings to a classifier layer for binary classification of the submissions. We emphasize that in this technique, we do \textit{not} utilize the submission content to demonstrate the post-to-post graph representativeness.

Table \ref{Table:p2p.result} provides the cross-validation results for the different models. Like the previous experiment, we divide the data into the train ($80\%$), validation ($10\%$), and test ($10\%$) sets. All models undergo a validation process, and we report the results for the test set. The combination of {CREDiBERT} with the Graph Convolutional Network (GCN) demonstrates superior performance compared to the baseline models. Specifically, the table highlights more than $10 \%$  improvement in the F1 score compared to the S-BERT model, illustrating the effectiveness of {CREDiBERT} when combined with graph-based analysis techniques.
The F1 score of $0.677$ achieved by utilizing only node2vec embeddings indicates that the post-to-post graph is proficient in encoding user reactions to detect low-credibility submissions. 

Overall, the results underscore the value of incorporating interaction information through a weighted post-to-post network into the model for enhanced credibility assessment. Lastly, we emphasize that due to the removal of the submission from {r/politics} and the smaller number of submissions in this experiment, the difficulty of the task has increased significantly, justifying the slight performance loss compared to the previous experiment.

\subsection{Topic-based Susceptibility Analysis}

To demonstrate the capabilities of the CREDiBERT, we conduct a case study on the susceptibility of different subreddit communities toward low-credible sources with respect to different topics. By analyzing the credibility of submissions in these communities, we aim to gauge their exposure to low credible information \cite{mosleh2022measuring}.
% \ege{Ege: I think it would be better to stick with either low credibility or misinformation. mixing them raises questions about whether they are the same thing }
 Reddit voting system, where users upvote or downvote submissions, offers insights into community responses to different information sources. We use submission scores, which reflect the average of upvotes and downvotes and the diverse submissions covering major events in each subreddit, to determine each community's susceptibility to specific topics. For this study, we identify the major topics discussed in the submissions for all the subreddits (verified and unverified) posted during January $2022$ in five major subreddits and utilize the BERTopic model, a method developed by Grootendorst \citeyear{grootendorst2022bertopic} for topic modeling.

Let us consider topic $h$ and the set of submissions referencing $h$ posted in subreddit $z$ by $\mathcal{S}_{h,z}$. We  measure the exposure of users in subreddit $z$ to low credible information, $\gamma_{h,z}$, by averaging the credibility label of the  submissions given by
\begin{equation}
    \gamma_{h,z} = \frac{1}{|\mathcal{S}_{h,z}|} \sum_{s\in \mathcal{S}_{h,z}} \gamma_s,
\end{equation}
where $|\mathcal{S}_{h,z}|$ denote the number of submissions addressing topic $h$ in subreddit $z$, and $\gamma_s \in \{0,1\}$ stands for binary credibility label for submission $s$ generated through CREDiBERT.
To estimate the reaction of the subreddit user to topic $h$, we calculate the weighted average of the credibility labels of submission in $\mathcal{S}_{h,z}$ with respect to the submission score, given by
\begin{equation}
    \rho_{h,z} = \frac{\sum_{s\in \mathcal{S}_{h,z}} \iota_s \lambda_s}{\sum_{s\in \mathcal{S}_{h,z}} \lambda_s},
\end{equation}
where $ \lambda_s$, and $\rho_{h,z} $ denotes the submission score and reaction score. The reaction score, $\rho_{h,z} $, reflects the credibility of the sources users promoted regarding the topic in question.

\begin{figure*}
	\centering
        \includegraphics[trim={0 0 0 0},clip,width=1\linewidth]{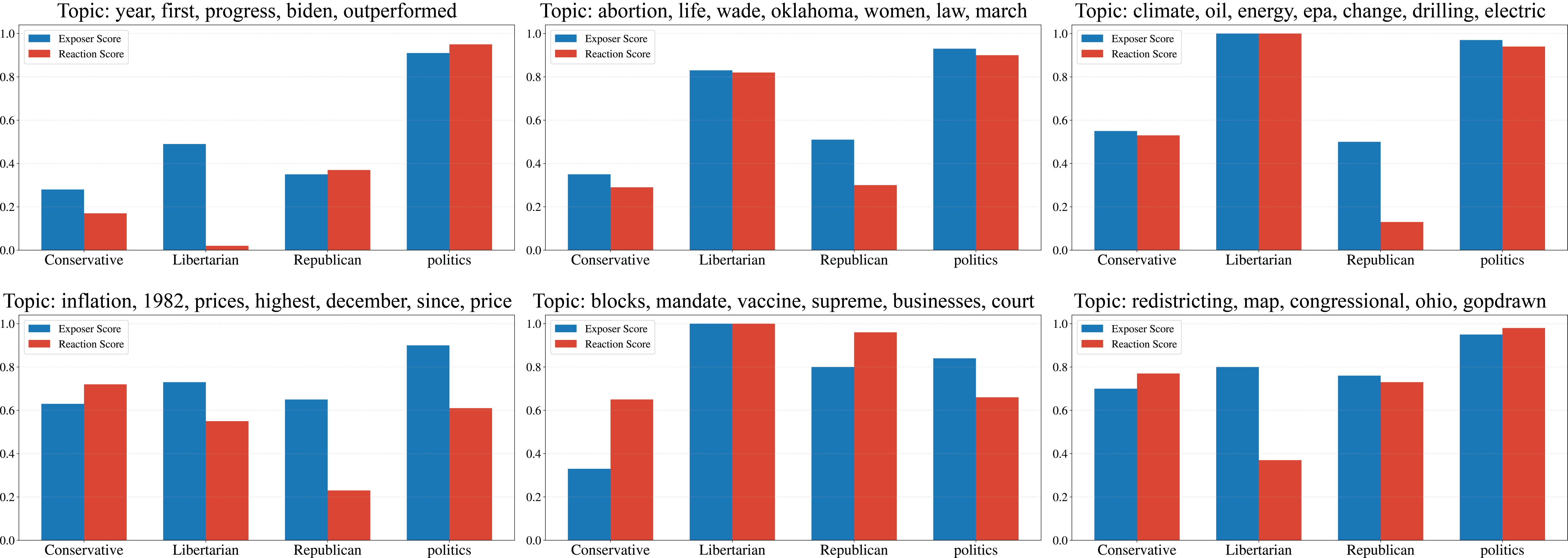}
	% \begin{subfigure}{.32\linewidth}
	% 	\includegraphics[trim={0 0 0 0},clip,width=\linewidth]{Figures/Topic-Biden2.png}
	% 	% \caption{\text{r/Conservative}}
	% 	% \label{fig:subfigA}
	% \end{subfigure}
	% \begin{subfigure}{.32\linewidth}
	% 	\includegraphics[trim={0 0 0 0},clip,width=\linewidth]{Figures/Topic-Energy2.png}
	% 	% \caption{\text{r/politics}}
	% 	% \label{fig:subfigB}
	% \end{subfigure}
 %        \begin{subfigure}{.32\linewidth}
	% 	\includegraphics[trim={0 0 0 0},clip,width=\linewidth]{Figures/Topic-abortion2.png}
	% 	% \caption{\text{r/Conservative}}
	% 	% \label{fig:subfigA}
	% \end{subfigure}
	% \begin{subfigure}{.32\linewidth}
	% 	\includegraphics[trim={0 0 0 0},clip,width=\linewidth]{Figures/Topic-Inflation2.png}
 %        \end{subfigure}
	% \begin{subfigure}{.32\linewidth}
	% 	\includegraphics[trim={0 0 0 0},clip,width=\linewidth]{Figures/Topic-Covid2.png}
 %        \end{subfigure}
	% \begin{subfigure}{.32\linewidth}
	% 	\includegraphics[trim={0 0 0 0},clip,width=\linewidth]{Figures/Topic-Redistirct2.png}
 %        \end{subfigure}
    \caption{The exposure (blue) and reaction (red) score of 6 topics in r/Conservative, r/Republican, r/Libertarian, and r/politics. Among all subreddits, r/politics has the highest exposure score, while r/Conservative and r/Republican have the lowest exposure score. While r/Libertarian shows extreme susceptibility to certain topics for others, it has identical exposure and reaction scores.}
    \label{fig:Exposure}
\end{figure*}

Figure \ref{fig:Exposure}  illustrates the exposure and reaction scores for six topics across four subreddits: r/Conservative, r/Republican, r/Libertarian, and r/politics. We select topics with a minimum of four submissions in each subreddit and ensure a balanced representation across the political spectrum. To assess a community susceptibility to misinformation, we consider two factors: the reaction score $\rho_{h,z}$, which reflects the credibility of sources favored by users, and the exposure score $\gamma_{h,z}$, reflecting the sources available to the community, predominantly controlled by subreddit moderators. A lower exposure score than a reaction score in a subreddit suggests a preference for more credible sources than those provided by moderators. Therefore, the reaction score ($\rho_{z,h}$) and the difference between the reaction and exposure scores ($\rho_{z,h}-\gamma_{z,h}$) are crucial indicators of a community susceptibility to specific topics.

 We observe that subreddits generally identified as right-leaning, such as r/Conservative, show lower exposure scores than those identified as left-leaning. Specifically, in r/Conservative, there is a notable trend: users tend to promote submissions from more credible sources when the topics align with right-leaning perspectives (second row of Figure \ref{fig:Exposure}). Conversely, users favor less credible sources more frequently when topics contradict their biases. This pattern highlights the influence of political biases on the selection of information sources within these online communities. However, in order to arrive at any concrete conclusions, conducting a more comprehensive study is necessary.

Subreddit r/politics shows the highest exposure to credible information among the subreddits in this study. However, we observe a trend where users chose to promote sources with lower credibility in discussions on topics like `\textit{Highest inflation rate since 1982}' and `\textit{Supreme Court ruling on vaccine mandate}'. This suggests that, for certain topics, the community's exposure to credible sources does not necessarily prevent the selection of less reliable information. This phenomenon indicates a susceptibility within r/politics to favor less credible sources when discussing specific, perhaps more controversial, topics. While these observations from r/politics and r/conservative do not establish a direct correlation with susceptibility to fake news, they suggest a potential for greater susceptibility to these topics compared to other communities. This is particularly notable in cases where users, despite being exposed to low-credibility sources, predominantly vote in favor of more credible ones.

% \vspace{-.2mm}

\section{Discussion}
CREDiBERT has shown notable efficacy in identifying low-credibility submissions on Reddit, outperforming conventional text classification methods. While currently optimized for Reddit data, its design holds potential for adaptation to other social media platforms and news websites, broadening its applicability. Furthermore, the automated pairing of submissions in CREDiBERT facilitates ongoing model refinement, enabling it to stay current with evolving trends and patterns in online misinformation.
% Moreover, the model holds the potential for automating the detection of high-risk posts on social media, which could be pivotal in combating misinformation. 
Our findings indicate that even in communities with access to reliable sources, there is a tendency to favor less credible information for certain topics. This insight opens up avenues for using CREDiBERT to understand and potentially mitigate topic-specific misinformation.

The weighted post-to-post network offers a novel method to analyze user interactions without delving into personal user data. This method encodes extensive user interactions in a weighted graph, enabling comprehensive analysis while respecting user privacy. While platforms like Twitter and Facebook present different challenges due to their content structure, adapting the post-to-post methodology could be a promising path forward.
 
% \paragraph{Limitations}
\textbf{Limitations:} A major limitation of CREDiBERT lies in its ability to assess only the credibility of news sources rather than the veracity of the content within the articles. This means that while it can effectively identify patterns of misinformation from generally unreliable sources over time, it may not detect false information disseminated through otherwise credible outlets. This gap highlights the challenge of discerning nuanced misinformation.

Unreliable news sources often attempt to mimic the style of credible ones, particularly in their headlines. This presents a second challenge for CREDiBERT: as these distinctions blur over time, text classification can struggle to differentiate between them. However, incorporating a post-to-post network could potentially mitigate this issue. By analyzing users reactions, we can glean additional insights into the credibility of posts, as demonstrated earlier.
% \vspace{-.2mm}
\section{Conclusion}

In this study, we introduced CREDiBERT, an innovative sentence-level embedding model designed to assess credibility in Reddit submissions. CREDiBERT outperforms existing text classification models for this task, according to our evaluations. We also developed a weighted post-to-post network to encode Reddit user interactions efficiently without requiring user profiling. When integrated with CREDiBERT, this network enhances the detection of credible sources. By applying CREDiBERT to recent Reddit submissions, we have revealed its capability to estimate community susceptibility to low-credible information on various topics. Exploring the application of CREDiBERT in other social media contexts and refining its methodology to address its current limitations present exciting avenues for future research.

\section{Ethics Statement}

The research presented in this paper strictly utilizes publicly available data from Reddit. In line with ethical standards, our methodology prioritizes user anonymity; we do not track or profile any users. The CREDiBERT framework is designed to analyze user interactions while fully respecting user privacy, avoiding the inclusion of any personally identifiable information. We recognize the potential for CREDiBERT to be used unethically. For example, sources with low credibility might attempt to leverage our model to modify their content, making it appear more credible. To mitigate such risks, we advocate for strict ethical guidelines governing the use of CREDiBERT and similar tools.
% We strongly encourage researchers and practitioners to adhere to these guidelines, ensuring that these technologies are used responsibly and ethically, without compromising the integrity of information sources.

\bibliography{Ref.bib}

\end{document}